\documentclass[conference]{IEEEtran}
\usepackage{cite}
\usepackage{booktabs}
\usepackage{multirow} 
\usepackage[pdftex]{graphicx}
\graphicspath{{../figures/}}
\DeclareGraphicsExtensions{.pdf,.jpeg,.png}
\usepackage[cmex10]{amsmath}
\usepackage{array}
\usepackage{mdwmath}
\usepackage{mdwtab}
\usepackage[font=footnotesize,caption=false]{subfig}
\usepackage{url}

\begin{document}

%
\title{Weakly Supervised Food Image Segmentation using \\ Vision
  Transformers and Segment Anything Model}

\author{\IEEEauthorblockN{Ioannis Sarafis, Alexandros Papadopoulos and Anastasios Delopoulos}
\IEEEauthorblockA{%
Department of Electrical and Computer Engineering\\
Aristotle University of Thessaloniki, Greece\\
Email: sarafis@mug.ee.auth.gr, alpapado@mug.ee.auth.gr, antelopo@ece.auth.gr}}

\maketitle

\begin{abstract}
In this paper, we propose a weakly supervised semantic segmentation
approach for food images which takes advantage of the zero-shot
capabilities and promptability of the Segment Anything Model (SAM)
along with the attention mechanisms of Vision Transformers
(ViTs). Specifically, we use class activation maps (CAMs) from ViTs to
generate prompts for SAM, resulting in masks suitable for food image
segmentation. The ViT model, a Swin Transformer, is trained
exclusively using image-level annotations, eliminating the need for
pixel-level annotations during training. Additionally, to enhance the
quality of the SAM-generated masks, we examine the use of image
preprocessing techniques in combination with single-mask and
multi-mask SAM generation strategies. The methodology is evaluated on
the FoodSeg103 dataset, generating an average of 2.4 masks per image
(excluding background), and achieving an mIoU of 0.54 for the
multi-mask scenario. We envision the proposed approach as a tool to
accelerate food image annotation tasks or as an integrated component
in food and nutrition tracking applications.
\end{abstract}

\begin{figure*}[th!]
  \centering
  \centering
  \includegraphics[width=0.98\linewidth]{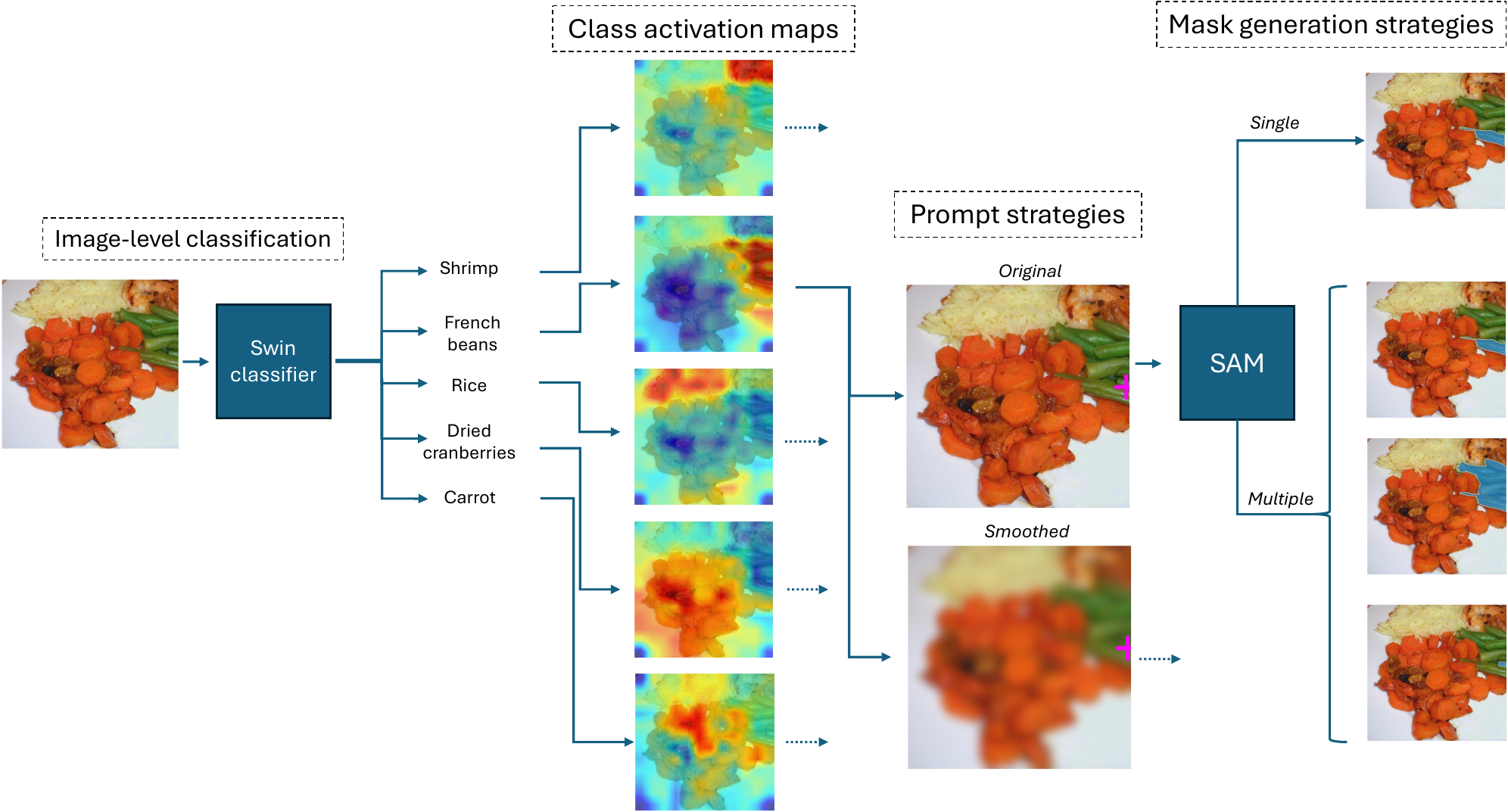}
  \caption{Overview of the proposed methodology using an example image
    for illustration. For clarity in the diagram, similar or repeated
    paths are omitted and represented using dashed arrows. First, the
    input image is passed through a Swin Transformer model that has
    been fine-tuned for food image-level classification. Then, for
    each predicted class, we apply Grad-CAM algorithm to compute the
    corresponding class activation map (CAM). The point with the
    highest activation value in each CAM is selected as a prompt for
    Segment Anything Model (SAM). The original or a smoothed version
    of the image can be used as input to SAM depending on our prompt
    strategy, with each option having certain trade-offs. Finally, SAM
    can be used to either produce a single mask or multiple masks,
    with the latter having significant performance gains in
    semi-automatic usage scenarios.}
  \label{figure:overview}
\end{figure*}

\section{Introduction}
The demand to accurately detect and quantify food using computer
vision methods is ever-increasing, driven by the demands of modern
approaches for digital monitoring of diet and personalized
interventions. Nevertheless, the algorithmic advancements in image
segmentation have outpaced the availability of food image datasets
with pixel-level annotations \cite{min2023large}, which can be
attributed to the labor-intensive task of producing accurate mask
annotations.

In our previous work \cite{vlachopoulou2023food}, we demonstrated that
{\bf Weakly Supervised Semantic Segmentation (WSSS)} methodologies can
be a feasible approach for building semantic food image segmentation
models. The models featured a patch-based attention mechanism and were
trained solely with image-level annotations for image classification
tasks. The attention mechanism was used to simultaneously detect the
class of the image and generate a class heatmap. Thresholding the
heatmap would generate the segmentation mask for the food class.
While our previous approach showcased the viability of WSSS methods
for this task, it also had significant limitations. Most notably, it
was necessary to train separate binary models to take advantage of the
heatmaps for distinct food classes, limiting scalability.

Since then, advances in zero-shot image segmentation have opened the
way for new approaches. The {\bf Segment Anything Model (SAM)}
\cite{kirillov2023segment} was designed for the \emph{promptable
segmentation task} which aims to generate a valid image segment for
any given prompt, such as points and boxes. As a foundational model,
SAM excels in general-purpose zero-shot segmentation but it can
struggle in domain-specific segmentation tasks with
out-of-distribution data \cite{li2024segment,tabassum2024adapting}.

In this work, we demonstrate how {\bf Vision Transformers (ViTs)} can
be used for food image classification while simultaneously leveraging
the resulting CAMs with SAM to generate segmentation masks for the
predicted classes. ViTs, inspired by the Transformer architecture
widely used in natural language processing (NLP), were introduced by
Dosovitskiy et al. \cite{dosovitskiy2020image} as an alternative to
convolutional neural networks (CNNs), outperforming the latter in many
tasks. ViTs replace the convolutional layers with a self-attention
mechanism applied to image patches. Our proposed methodology takes
advantage of the self-attention capabilities of ViTs to generate
precise query prompts for SAM.

The key contributions of this work are:

\begin{itemize}

\item Introduction of a food image segmentation approach relying solely
  on the image-level labels for training, eliminating the need for
  pixel-level annotations.

\item Examination of an image preprocessing technique using Gaussian
  blurring, which can enhance the quality of final SAM-generated masks in
  certain usage scenarios.

\item Evaluation of single-mask and multi-mask strategies, with the
  latter leading to improved segmentation performance for
  semi-automatic usage scenarios.

\end{itemize}

The remainder of the paper is organized as follows. First, Section
\ref{section:related_work} gives a brief overview of the related
work. Section \ref{section:methodology} describes the proposed
methodology. Then, Section \ref{section:experiments} presents the
experiments and results. Finally, Section \ref{section:conclusions}
concludes with a discussion on the results and the potential
directions for future work.

\section{Related work}
\label{section:related_work}
A first attempt to exploit SAM for food image classification was made
by Lan et al. \cite{lan2023foodsam}, though the role of SAM was
limited solely to refining the masks generated by a SETR
\cite{zheng2021rethinking} segmentation model that had been trained on
pixel-level annotations. While this approach led to enhancements by
improving the SETR masks using the SAM-generated masks, it still
suffered from the shortcoming of requiring pixel-level annotations to
train the initial segmentation model. Chen et
al. \cite{chen2024ingredsam} proposed a method that combines feature
maps from multiple visual foundation models to produce prompts for
SAM. Their method uses clean images with a single ingredient to
produce a feature map, which is then compared with the feature map of
the image to be segmented in order to produce the prompts. Thus, they
are able to take advantage of the SAM capabilities for zero-shot
segmentation with a limitation of this method being that the target
class must be known in order to apply the appropriate image
queries. To directly produce segmentation masks, Alahmari et
al. \cite{alahmari2024segment} fine-tuned SAM using Low-Rank
Adaptation layers; however, the fine-tuned model is capable of
producing only binary masks that separate food from the background.

When using image classifiers, the class activation maps (CAMs) can be
used as a starting point for WSSS problems \cite{li2024area}. In the
context of medical images, Wang et al. \cite{wang2025weakmedsam}
combined the prompt capability of SAM to improve the segments
generated by CAMs. Similarly, Kweon and Yoon \cite{kweon2024from}
proposed the use of CAMs obtained by the classifier to generate
prompts for SAM that result in class-wise SAM masks. Their approach
was evaluated on semantic segmentation of objects in natural
images. In this work, we also adopt the usage of CAMs with SAM
prompting and demonstrate our approach's viability for food image
segmentation.

\section{Proposed Methodology}
\label{section:methodology}
The methodology consists of an image-level classification stage,
followed by a zero-shot segmentation stage, described in detail
below. Figure \ref{figure:overview} illustrates the proposed approach.

\subsection{Image-level food classification}
\label{subsection:training}
We begin by training an image classifier using image-level
labels. This is the only component in the proposed methodology
requiring training. For this purpose, we selected the Swin Transformer
\cite{liu2021swin}.

The Swin Transformer constructs hierarchical feature maps using a
shifted window mechanism to compute self-attention. This architecture
achieves high performance in classification tasks and produces feature
maps that capture information at both fine and broader scales, making
it suitable for semantic segmentation tasks as well. Additionally, the
Swin Transformer has shown that it can be effectively fine-tuned on
downstream tasks with smaller datasets \cite{xu2023swin}. The latter
is a significant advantage for this architecture, allowing us to
replace the original attention mechanism from our previous work
\cite{vlachopoulou2023food} with a more powerful model that
demonstrates strong generalization performance.

As multiple food classes typically coexist within an image, we
fine-tune the model using a multi-label classification setting for $N
+ 1$ classes, where $N$ is the number of food classes, and an extra
class is used for the background. We apply sigmoid activation
$\sigma(\cdot)$ on the output layer, and use the binary cross-entropy
loss during training:
\begin{equation}
 l(x,y) = -\frac{1}{N+1}\sum_{i=1}^{N+1}[y_i \log(\hat{y}_i) + (1 - y_i)\log(1 - \hat{y}_i)]
\end{equation}
where $y_i$ is the ground-truth label (0 or 1) for class $i$, and
$\hat{y}_i$ is the predicted probability for class $i$ (i.e., the
output of the sigmoid).

\subsection{Zero-shot food segmentation}
Given a new image, we first apply the fine-tuned classifier to predict
the food classes within it. Then, for each predicted class, we apply
the Grad-CAM algorithm \cite{jacobgilpytorchcam} to compute the
corresponding class activation map (CAM). Specifically, we apply
Grad-CAM to the output of the final normalization layer of the Swin
Transformer which lies after all transformer blocks and right before
the classification head.  This layer captures high-level semantic
information for the image regions that are most relevant to the
predicted class.

Having constructed the CAMs, we select points to act as queries for
prompts to SAM. In this work, we follow a simple approach where we
select the point corresponding to the highest activation value within
each CAM. We examine two variations for the input image: (a) using the
original image; or (b) a smoothed image that has been preprocessed
using a Gaussian blur filter. The rationale behind the Gaussian blur
is to produce more unified regions for the same food item, helping SAM
bridge visual boundaries between adjacent elements of the same food
type (e.g., carrot pieces).

Lastly, regarding the output masks, we evaluate the capability of SAM
to produce for a prompt: (a) a single mask by averaging multiple valid
masks; and (b) multiple valid masks. The latter can lead to
significant performance gains with little user interaction. It
simulates a usage scenario in a semi-automatic setting where an
annotator/user chooses the most accurate mask from a small set of
candidate ones.

\begin{figure*}[!htbp]
  \centering
  \includegraphics[width=0.99\linewidth]{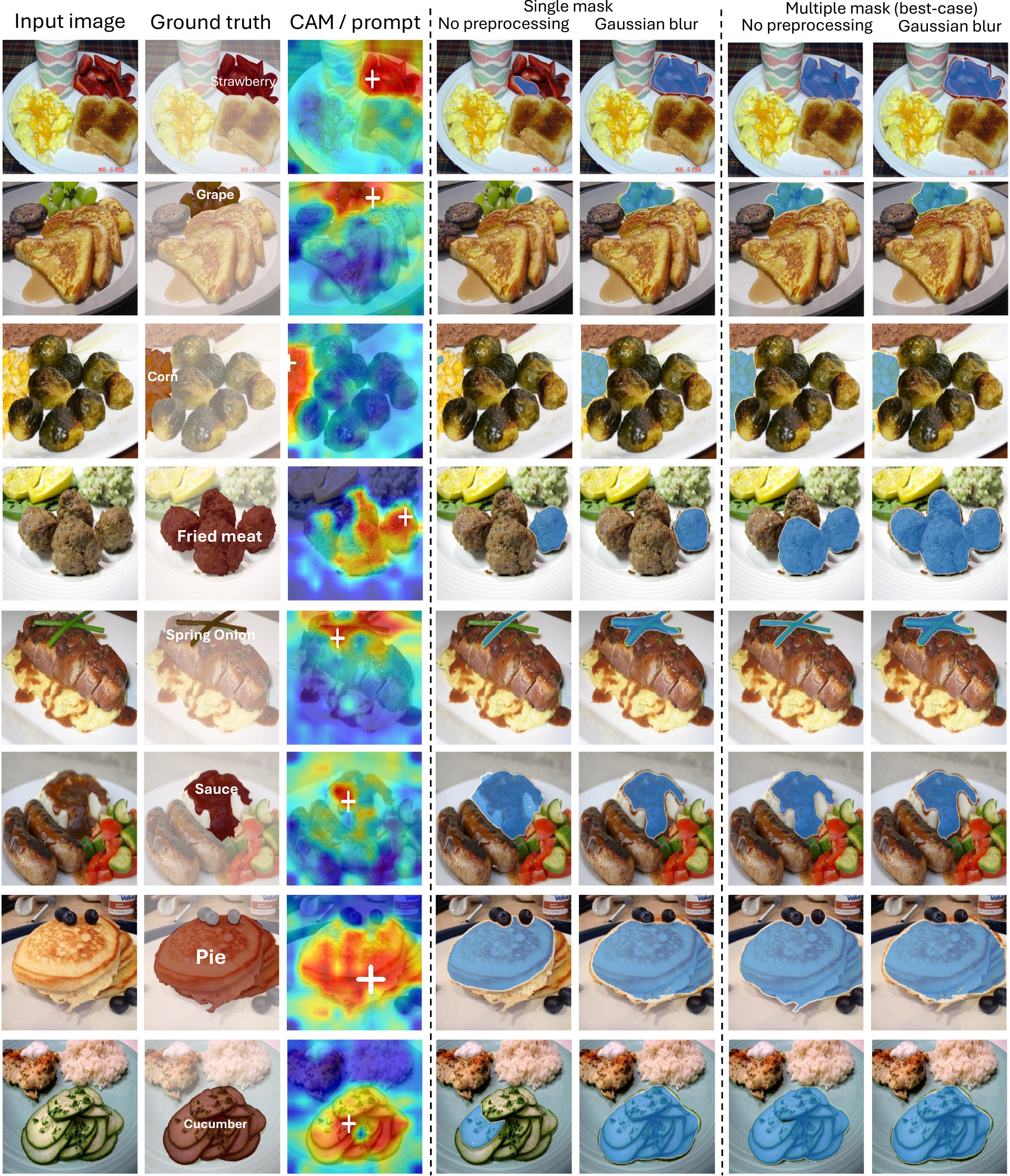}
\caption{Examples of segmentation masks generated using the proposed
  method for various food classes within images of the test set. From
  left to right, the columns show: (1) the original input image, (2)
  the ground truth segmentation mask for the class, (3) the computed
  Class Activation Map (CAM), with a cross marking the point of
  highest activation used as the prompt, (4) the segmentation mask
  produced by SAM using the single-mask strategy with the original
  image, (5) the corresponding single-mask output using a smoothed
  version of the image, (6) the best-performing mask among the top-3
  SAM outputs (multi-mask strategy) using the original image, and (7)
  the corresponding best mask when using multi-mask strategy and the
  smoothed image.}
  \label{figure:examples}
\end{figure*}

\begin{table*}[htbp]
\centering
\caption{Segmentation performance comparison for different
  preprocessing methods and SAM mask strategies. \\ Best performance for each
  column is marked in bold.} %
\label{table:results}
\resizebox{\textwidth}{!}{
\begin{tabular}{l c c c c c c c c c c c c c}
\toprule
\multirow{2}{*}{\textbf{Input image}} & 
\multirow{2}{*}{\textbf{\# Mask strategy}} &
\multirow{2}{*}{\textbf{mIoU}} & 
\multicolumn{10}{c}{\textbf{IoU for ten most frequent classes}} \\
\cmidrule(lr){4-13}
& & & Bread & Carrot & Chicken/Duck & Tomato & Steak & Sauce & Broccoli & Potato & Ice cream & Cilantro mint \\
\midrule
Original (no preprocessing) & Single & 0.29 & 0.39 & 0.13 & 0.36 & 0.23 & 0.35 & 0.41 & 0.33 & 0.51 & 0.53 & 0.26 \\
Original (no preprocessing) & Multi & \bf 0.54 & \bf 0.53 & \bf 0.60 & \bf 0.64 & \bf 0.48 & \bf 0.57 & \bf 0.63 & \bf 0.77 &\bf 0.69 & \bf 0.61 & 0.38 \\
Smoothed (Gaussian blur)  & Single & 0.36 & 0.33 & 0.35 & 0.40 & 0.28 & 0.36 & 0.47 & 0.53 & 0.48 & 0.17 & 0.28 \\
Smoothed (Gaussian blur) & Multi & 0.51 & 0.47 & 0.58 & 0.60 & 0.47 & 0.50 & 0.56 & 0.75 & 0.64 & 0.20 & \bf 0.39 \\
\bottomrule
\end{tabular}}
\end{table*}

\section{Experiments}
\label{section:experiments}
We conduct experiments to evaluate the proposed method using the
FoodSeg103 dataset \cite{wu2021foodseg103}. The dataset contains $103$
food classes and is split into a training set with $4,983$ images and
a test set with $2,315$ images.

\subsection{Training configuration}
Both training and test set include pixel-level annotations for food
segments; however, we do not make use of them during training. For
this purpose, we aggregate the pixel-level class masks to simulate
a multi-label, image-level classification problem.

We use the large-sized Swin model (approx. $197$ million weights) that
was pre-trained on ImageNet-21k (approx. $14$ million images, $21,841$
classes) at resolution $384 \times 384$. The model divides images into
non-overlapping patches of size $4 \times 4$ pixels and uses a window
size of $12 \times 12$ pixels to compute self-attention.

To improve the robustness of the classifier, we apply an augmentation
strategy for training data that includes: (a) randomly resizing with
scale variations from $80\%$ to $100\%$ and then randomly cropping to
a size of $384 \times 384$; (b) performing random horizontal and random
vertical flips with $50\%$ probability; (c) color jittering ($\pm
20\%$ for brightness, $\pm 20\%$ for contrast, $\pm 15\%$ for
saturation, $\pm 10\%$ for hue); (d) random affine transformations
(rotations up to $30^\circ$, vertical and horizontal translations up to
$10\%$, scaling adjustments between $90\%$ and $110\%$); (e) Gaussian
blur using a kernel size of $3 \times 3$ and sigma between $0.001$ and
$2.0$; and (f) random erasing with probability $0.3$ of small regions
on the image (area ratios from $2\%$ to $10\%$ with aspect ratio
between $0.3$ and $3.3$).

As the dataset lacks a dedicated validation set, the model is
fine-tuned for a predetermined number of $50$ epochs, using a cosine
learning rate scheduler with a warmup phase for $10$ epochs. The
initial learning rate is set to $2 \times 10^{-4}$ and we apply
regularization with a weight decay of $10^{-4}$. The batch size is set
to $64$.

\subsection{SAM configuration}
We use SAM 2, specifically the SAM 2.1 checkpoint for the large model
(sam2.1\_hiera\_large), which contains approximately $222.4$ million
parameters \cite{ravi2024sam2}. Compared to the original Segment
Anything Model (SAM), SAM 2 is more accurate and faster for image
segmentation tasks.

When using a smoothed image for input to SAM, we apply a Gaussian
filter with $\sigma=10$ across the horizontal and vertical axes. In
the experiments with multi-mask proposals, we set their number to $3$.

\subsection{Evaluation metrics}
For the evaluation, we focus on the performance of the true positive
predictions by the classifier, since our primary goal is to assess the
quality of the masks for correctly detected classes. This aligns with
our envisioned use of the methodology as a tool to accelerate
annotations and reduce manual annotation effort for food segmentation
tasks.

We use the \emph{mean Intersection over Union (mIoU)} metric which is defined as:
\begin{equation}
    \text{mIoU} = \frac{1}{C}\sum_{c = 1}^{C} \frac{TP_c}{TP_c + FP_c + FN_c}
\end{equation}
where $C$ is the number of classes, $TP_c$ is the number of pixels
correctly identified as belonging to class $c$, $FP_c$ is the number
of pixels incorrectly predicted as belonging to class $c$, and $FN_c$
is the number of pixels belonging to class $c$ that were omitted by
the predicted masks.

When we opt for multiple SAM mask proposals, we report the best-case
IoU by selecting, for each class and for each image, the mask with the
highest IoU against the ground truth.

\subsection{Results}
Table \ref{table:results} shows the mIoU results calculated over
$5571$ mask proposals (excluding the background class). Separate IoU
scores for the $10$ most frequent classes in the evaluation set are
also included. Figure \ref{figure:examples} shows test set examples
alongside their corresponding CAMs with query points (marked with a
cross), and the resulting segmentation masks. Results are shown for
all combinations of input preprocessing (original and smoothed) and
mask proposal strategies (single-mask and multi-mask).

When using single-mask proposals, the best performance is achieved
with smoothed images, resulting in an mIoU of $0.36$. This indicates
that using more unified image regions can lead to improved
segmentation performance in food images. In comparison, the
single-mask proposal using the original image leads to an mIoU of
$0.29$. However, as expected, using the smoothed image also results in
coarser boundaries between adjacent regions.

On the other hand, when using multi-mask proposals, the best-case mIoU
of $0.54$ is obtained without any preprocessing, indicating that SAM
performs better on original images when there is the potential to
exploit several candidate masks per class in order to produce sharper
segmentation masks. Using the smoothed image with multi-mask proposals
reduces performance to $0.51$.

\section{Conclusions}
\label{section:conclusions}
In this work, we proposed a weakly supervised method for food image
segmentation that takes advantage of the attention mechanisms of ViTs
and their explainability using the Grad-CAM algorithm, in combination
with the strong zero-shot capabilities of SAM. Using a Swin
Transformer trained solely on image-level labels, we generate prompts
for SAM to produce segmentation masks, eliminating the need for
pixel-level ground truth during training.

Experiments conducted on the FoodSeg103 dataset demonstrate that the
proposed method can generate accurate segmentation masks with high
performance. Additionally, we verified that simple image preprocessing
techniques, such as Gaussian blur, can improve segmentation quality
when SAM is restricted to single-mask proposals. This approach is more
suitable when a rough estimate of the food is needed, e.g., within
nutrition tracking applications. We also evaluated the scenario where
SAM is used to propose multiple masks per prompt, achieving strong
best-case performance when original images are used. In practical
applications, this can be used within semi-automatic workflows where
an annotator or user selects the best mask from a small set of proposals.

A key limitation is that the image classifier's true positive rate
strongly influences the method's effectiveness as an annotation-
assisting tool, because correct predictions are required to generate
precise SAM prompts.

\subsection{Future work}
It is worth noting that our current work was constrained by the
relatively small size of the FoodSeg103 training set. As a next step,
we intend to leverage larger food image datasets with image-level
annotations to improve classification performance, potentially leading
to further improvements in overall performance.

Regarding the image classifier, we opted for the Swin Transformer
which has a proven track record on segmentation tasks. Alternative
models and a sensitivity analysis for the training parameters are left
for future work; these would clarify architecture trade-offs (e.g.,
accuracy vs. efficiency), and quantify the stability and
generalization of the proposed approach.

Finally, in this approach we used only the single highest-activation
point to prompt SAM. While this serves as a good baseline, more
sophisticated prompting strategies can be explored, such as using
multiple positive and negative points in the prompts, and/or bounding
boxes. Post-processing the predicted masks could further improve
performance by eliminating overlaps among food classes.

\section*{Acknowledgments}
This work has received funding from the Special Account for Research
Funds of the Aristotle University of Thessaloniki. This work is part
of a project that has received funding from the European Union’s
Horizon 2020 research and innovation programme under grant agreement
No. 965231.

\bibliographystyle{IEEEtran}
\bibliography{IEEEabrv,refs.bib}

\end{document}